\documentclass[letterpaper, 10 pt, conference]{IEEEconf}  

\IEEEoverridecommandlockouts                              

\usepackage{mathptmx} 
\usepackage{amsmath} 
\usepackage{amssymb}  
\usepackage{graphics} 
\usepackage{graphicx}
\usepackage[footnotesize]{subfigure}
\usepackage{subcaption}
\long\def\invis#1{}
\usepackage{cite}
\usepackage{url}
\usepackage{hyphenat}
\usepackage{amsmath,systeme}
\usepackage{xcolor}
\usepackage[linesnumbered,ruled]{algorithm2e}
\usepackage{siunitx}
\usepackage{enumerate}
\usepackage{mathtools}
\usepackage{booktabs}

\usepackage[utf8]{inputenc}
\usepackage{ wasysym }
\usepackage{pgfplots}
\usepackage{tikz}
\usepackage{gensymb}
\usepackage{pgfplotstable}
\usepgfplotslibrary{colormaps}
\pgfplotsset{colormap/hot}
\pgfplotsset{width=9cm,compat=1.9}
\pgfplotsset{layers/my layer set/.define layer set={background,main,foreground}{ },set layers=my layer set,}
\usepackage{xcolor}
\usepackage{tabularx}
\usepackage{multirow}
\usepackage{graphicx} 

\newcommand\fig[1]{Figure~\ref{#1}}

\newcommand\bmat{\begin{bmatrix}}
\newcommand\emat{\end{bmatrix}}
\usepackage{textcomp}
\usepackage{svg}
\usepackage{algpseudocode}
\usepackage{hyperref}
\usepackage{makecell}

\def\floatsep{4.0pt}
\def\textfloatsep{0.0pt}
\def\dblfloatsep{4.0pt}
\def\dbltextfloatsep{4.0pt}

\usepackage{fp}
\FPset{\pb}{0}
\newcommand{\pagebudget}[1]{}

\usepackage{xcolor}

\newif\ifmarkchange

\definecolor{changed}{rgb}{0.0,0.0,0.0}
\ifmarkchange
\definecolor{changed}{rgb}{0.91, 0.05, 0.05}
\fi

\title{
\LARGE \bf Underwater Dense Mapping with the First Compact 3D Sonar}
\author{Chinmay Burgul$^1$*, Yewei Huang$^2$*, Michalis Chatzispyrou$^1$, \\ \hspace{0.5cm}  Ioannis Rekleitis$^1$, Alberto Quattrini Li$^2$, and Marios Xanthidis$^3$
\thanks{$^1$ Chinmay Burgul, Michalis Chatzispyrou, and Ioannis Rekleitis are with  the Mechanical Engineering Department, University of Delaware, Newark, DE, USA. {\tt\small [cmburgul,michalis,yiannisr]@udel.edu}}
\thanks{$^2$ Yewei Huang and Alberto Quattrini Li are with Dartmouth College, Hanover, NH, USA {\tt\small \{yewei.huang, alberto.quattrini.li\}@dartmouth.edu}}
\thanks{$^3$ Marios Xanthidis is with the Aquaculture Robotics and Automation Group, SINTEF Ocean, {\tt\small marios.xanthidis@sintef.no}}
\thanks{* Authors contributed equally.}
\thanks{This work was supported by the Research Council of Norway (EchoNav: NO-359447), National Science Foundation (NSF 2545370, 1943205, 2024541, 2144624)}
}

\begin{document}

\maketitle
\thispagestyle{empty}
\pagestyle{empty}

\begin{abstract}
In the past decade, the adoption of compact 3D range sensors, such as LiDARs, has driven the developments of robust state-estimation pipelines, making them a standard sensor for aerial, ground, and space autonomy. 
Unfortunately, poor propagation of electromagnetic waves underwater, has limited the visibility-independent sensing options of underwater state-estimation to acoustic range sensors, which provide 2D information including, at-best, spatially ambiguous information. 
This paper, to the best of our knowledge, is the first study examining the performance, capacity, and opportunities arising from the recent introduction of the first compact 3D sonar. 
Towards that purpose, we introduce calibration procedures for extracting the extrinsics between the 3D sonar and a camera and we provide a study on acoustic response in different surfaces and materials.
Moreover,  we provide novel mapping and SLAM pipelines tested in deployments in underwater cave systems and other geometrically and acoustically challenging underwater environments.
Our assessment showcases the unique capacity of 3D sonars to capture consistent spatial information allowing for detailed reconstructions and localization in datasets expanding to hundreds of meters. 
At the same time it highlights remaining challenges related to acoustic propagation, as found also in other acoustic sensors.
Datasets collected for our evaluations would be released and shared with the community to enable further research advancements.
\end{abstract}


\section{Introduction}
\label{sec:intro}

Mapping underwater environments is a time consuming, labor intensive, and often dangerous operation. 
Overhead and cluttered environments are particularly challenging due to entanglement hazards. Underwater caves (and the inside of wrecks) present many dangers, even for highly skilled people, with more than $350$ deaths since $1969$ in USA alone~\cite{buzzacott2017recovery,casadesus2019diving}. 
However, Karst aquifers (underwater caves) account for 25\% of potable water worldwide~\cite{karstbook}. 
One of the main challenges in mapping underwater environments is the limited visibility. 
Even in clear waters, after a few meters, the scene becomes blurred and finding features is extremely challenging. 
Furthermore, limited illumination and color absorption~\cite{roznere2019iros,SkaffBMVC2008} result in most vision-based odometry/SLAM systems to under-perform~\cite{JoshiIROS2019}, which in turn affects autonomous pipelines that rely upon them~\cite{xanthidis2020navigation}.

\begin{figure}[t!]
    \centering
    \subfigure[]{%
        \includegraphics[width=0.48\linewidth, trim={0.0in, 0.0in, 0.0in, 0.17in},clip]{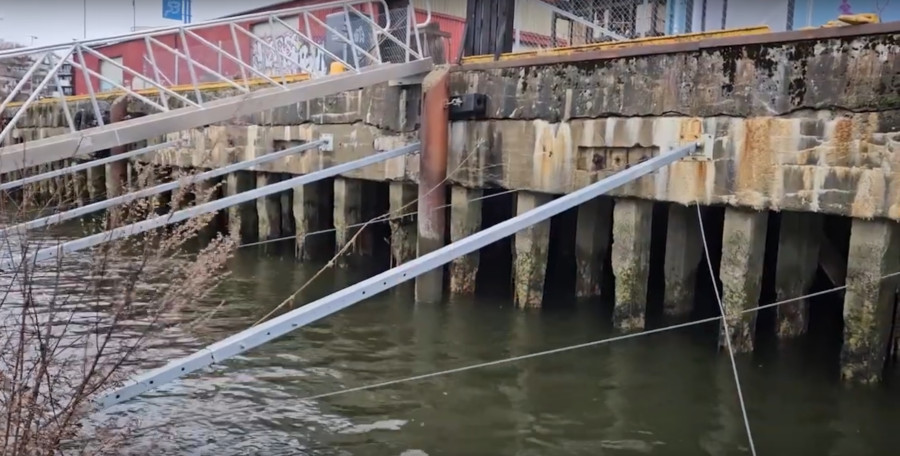}%
        \label{fig:murk:a}%
    }
    \subfigure[]{%
        \includegraphics[width=0.48\linewidth, trim={6.0in, 3.0in, 0.0in, 1.0in},clip]{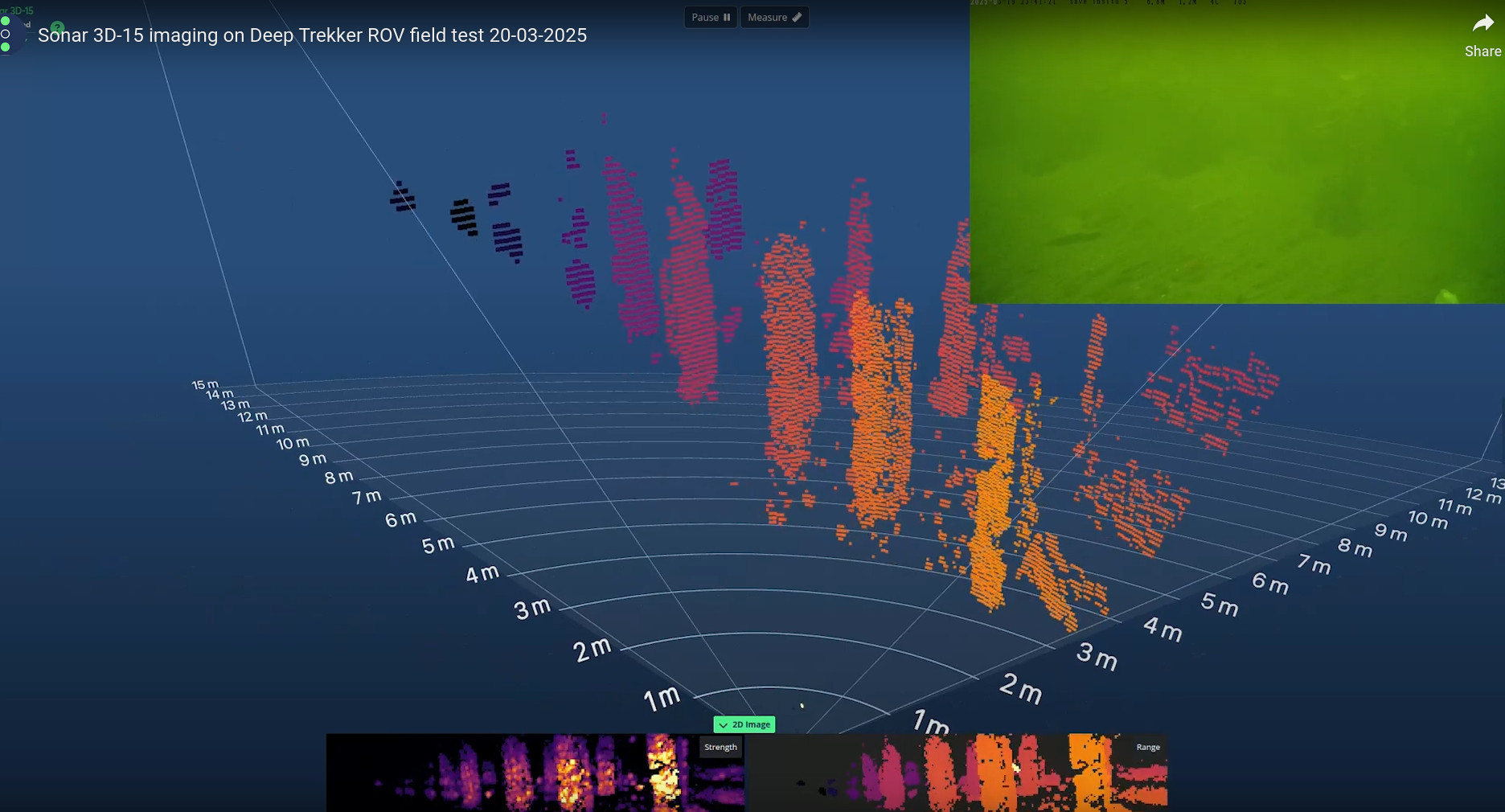}%
        \label{fig:murk:b}%
    }
    \subfigure[]{%
        \includegraphics[width=\linewidth,height=0.95\linewidth,keepaspectratio]{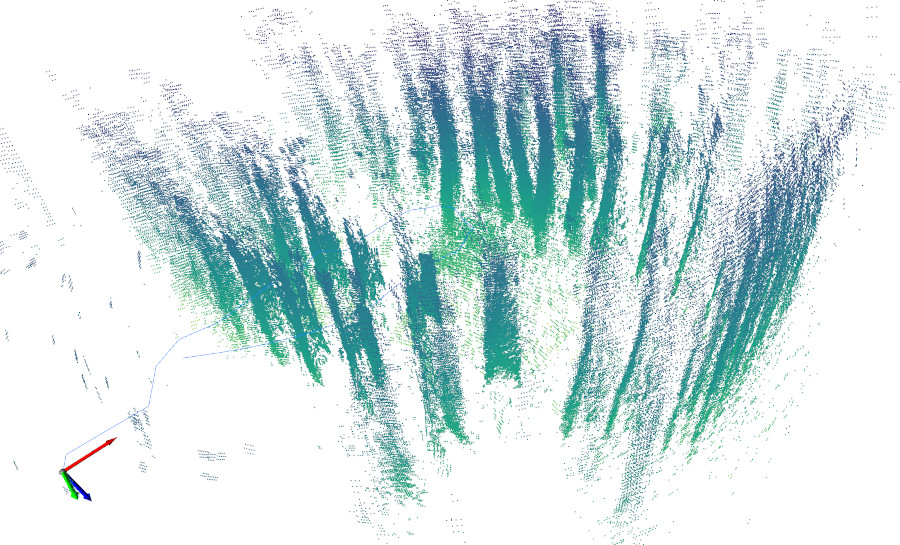}%
        \label{fig:murk:c}%
    }

    \caption{\textbf{3D sonar map reconstruction} in a pier piling environment with limited camera visibility. (a) Pier piling environment, (b) single 3D sonar scan and camera visibility, and (c) reconstructed pointcloud results.}
    \label{fig:murk}
\end{figure}

On the other hand, acoustic sensors --- despite their popularity --- suffer from several inherent limitations.
A primary drawback is their high angular uncertainty; for instance, the Ping360 from Blue Robotics~\cite{ping360}, a 2D mechanical scanning sonar, has a beam angle of $25^{\circ}$, which introduces substantial directional ambiguity.
Reducing this uncertainty requires narrower beams, but this comes at the expense of range. 
For example, the Imagenex pipe profiling sonar~\cite{imagenex}, another mechanical scanning sonar, achieves a beam width of only $1.4^{\circ}$, yet its effective range is limited to about 6 meters.
3D sonars may resolve some of these disadvantages, as showcased in preliminary works utilizing probabilistic scan matching~\cite{ferreira20223dupic}. 
Though, logistic challenges in these devices in terms of power demands, size, and weight hindered broad introduction to underwater platforms, limiting data collection to fly-overs in open-water environments.

In contrast, newly introduced 3D sonars, such as the Sonar~3D-15 from Water Linked~\cite{waterlinked}, are compact, provide a LiDAR-like pointcloud (as shown in Fig.~\ref{fig:murk}), and do not suffer from the aforementioned disadvantages, allowing for data collections in complex cluttered overhead environments.
Notably, this sensor can be readily integrated into most popular underwater platforms, providing acoustic measurements offering unique 3D spatial information.

This paper exploits the advantages of the Sonar~3D-15 with deployments spanning several field trials over different environments including a low-visibility dock, a submerged container, a swimming pool for testing the reflectance properties of different materials,  a cavern, 
and a \SI{300}{m} penetration of a cave system. 
With the exception of the dock and the container, \invis{see \fig{fig:doc}} visual, inertial, and water depth measurements were collected at the same time. 
This enabled the development of new calibration procedures, the introduction of SLAM frameworks that integrate such sensors, the collection of dedicated datasets, and the characterization of this novel 3D range sensor. 

In summary, the contributions of this paper are as follows:
\begin{enumerate}
    \item Characterization of the Sonar 3D-15 sensor's performance across different material types and surface properties.
    \item A novel calibration strategy for estimating the extrinsic parameters between the camera and sonar.
    \item A loosely coupled integration of sonar and VIO into a common estimation process, showing their complementary sensing capabilities.
    \item Introduction of several datasets, including sonar, visual, inertial, and water depth data, which would be made publicly available upon acceptance.
\end{enumerate}


\section{Related Work}
\label{sec:rel}

%
Simultaneous Localization and mapping (SLAM) are key to enabling autonomy in underwater environments, providing the foundation for reliable control and planning.
A wide range of sensor configurations have been explored, each tailored to the specific challenges of different environments and tasks. 
Johannsson et al. \cite{johannsson2010imaging} utilized imaging sonar and Doppler Velocity Logs (DVL) for harbor surveillance.
The Sunfish team~\cite{richmond2018OCEANS, sunfish} integrated a multibeam sonar, fiber-optic gyroscope (FOG), Inertial Measurement Unit (IMU), DVL and pressure sensors for cave exploration and mapping.
SVIn2 \cite{RahmanIJRR2022} combined visual-inertial sensing with a pipe profiling sonar, and pressure sensor data, extending the framework of OKVIS~\cite{leutenegger2015keyframe}, to map underwater caverns and shipwrecks; this approach was later extended to incorporate magnetometer data in~\cite{JoshiICRA2024}.
Turtlmap \cite{song2024turtlmapIROS} relied on DVL, inertial, and pressure sensors for localization in low-texture environments and a depth camera for mapping, while~\cite{zhang2024IET} utilized side-scan sonar (SSS) with dead-reckoning to address localization in open-water settings.

While proprioceptive sensors such as IMUs, DVLs, and magnetometers provide valuable localization cues, they lack environmental awareness for mapping. 
Vision-based approaches are constrained to short operating ranges due to light attenuation and turbidity in underwater environments, whereas traditional sonar-based methods, although resilient to visibility conditions, introduce their own challenges~\cite{mcconnell2022perception}.
Low-cost scanning sonars, such as Ping360 and pipe-profiling sonars, exhibit angular uncertainty and low-dimensional sensing.
Imaging sonar as well as side-scan sonars generate raw acoustic imagery that is subject to geometric distortion and uneven ensonification, causing the same region to appear significantly different when viewed from varying perspectives.
%
The Sonar 3D-15 extends these capabilities with a range of up to 15 meters. 
Unlike acoustic imaging sensors, its pointcloud is free from geometric distortion or vertical uncertainty, although it often displays uneven ensonification on smooth surfaces.
In this work, we evaluate a sonar-only approach in low visibility conditions and additionally explore fusion with Visual-Inertial sensors in environments with satisfactory visibility.
%

Calibrating the Sonar 3D-15 extrinsic parameters with a visual sensor appears conceptually similar to LiDAR-camera calibration.  
Existing approaches are broadly categorized into \emph{target-based} and \emph{targetless} methods.
In camera-pointcloud target-based calibration, past work~\cite{zhang2024cooperative} reviews the use of variety of targets, including 2D ArUco markers, checkerboards, and 3D objects (e.g., boxes, spheres, or other structured geometries).
Targetless methods exploit scene geometry such as structural regularities~\cite{nagyMVA19} or line-plane correspondences ~\cite{ZhouIros2018}. 
However, conventional flat or smooth calibration target surfaces are not reliably perceived by the tested 3D sonar, and target-less methods, relying on line-plane or geometric structure, are generally absent in natural underwater environments.
Morever, sonar sensors suffer form acoustic reverberation and acoustic multipath inference in confined environments (e.g., swimming pools), where hard, flat pool walls reflect acoustic pulses creating delayed echoes that return to the sensor from different angles at different times. 
This interference overwhelms the sensor's ability to distinguish a desired target due to the presence of such noise.
Similar practical challenges arise in using IMU-LiDAR calibration~\cite{chen2025troikalibr}, which additionally requires specific excitation patterns that are difficult to perform underwater.   
This motivated us to develop our own 3D sonar-camera calibration approach, which is presented in Section~\ref{subsec:calib}.

\begin{figure}[h]
    \centering
    \subfigure[]{%
        \includegraphics[width=0.32\columnwidth]{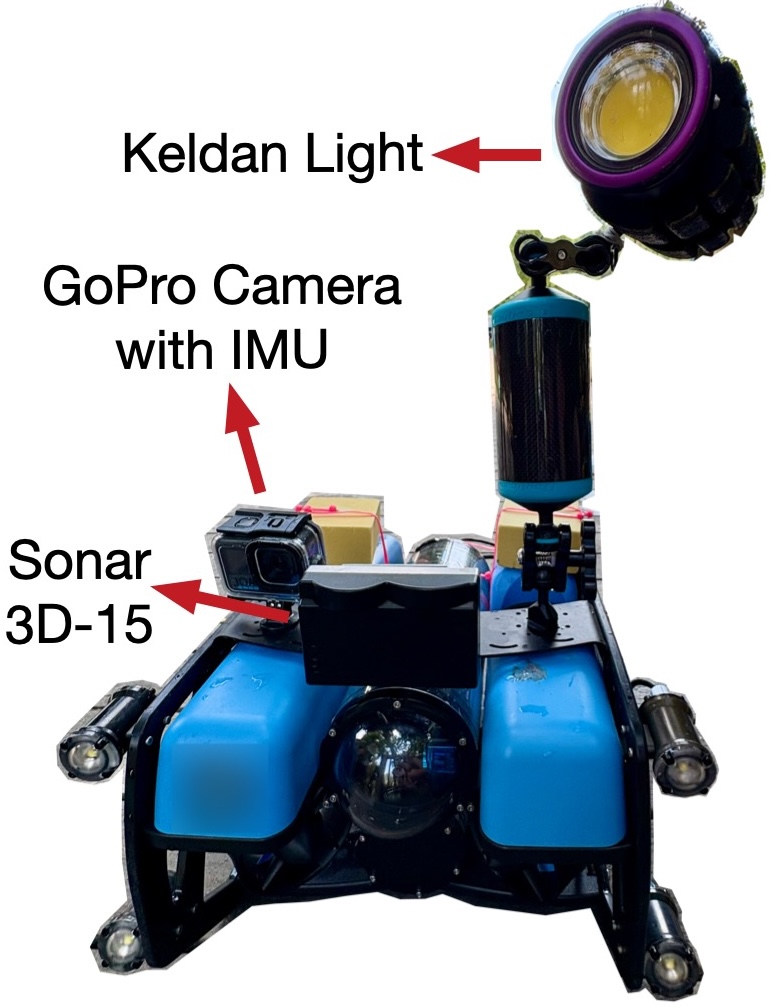}%
        \label{fig:beauty:a}%
    }
    \subfigure[]{%
        \includegraphics[width=0.65\columnwidth]{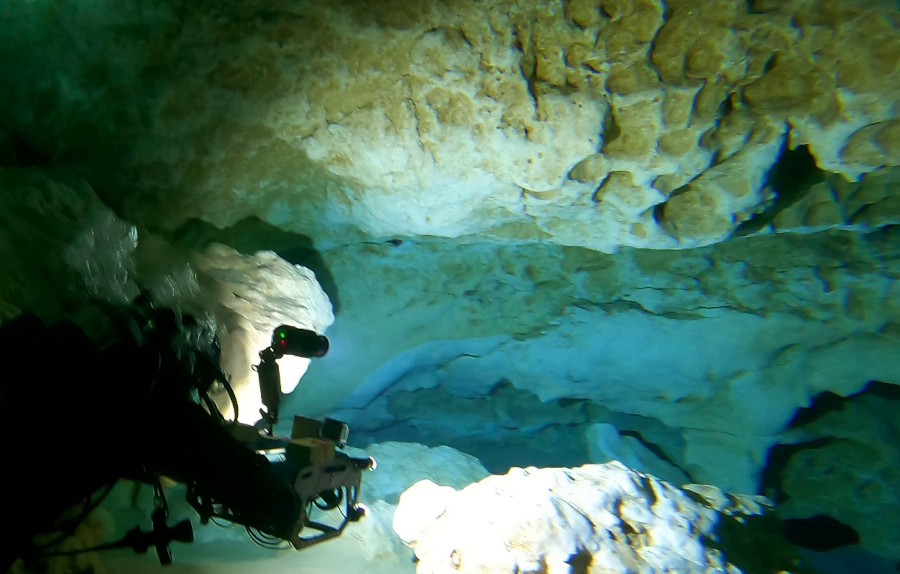}%
        \label{fig:beauty:b}%
    }
    \caption{Our customized BlueROV2 setup (a) and the experiment environment (b).}
    \label{fig:beauty}
\end{figure}

\section{Experimental Setup and Analysis}
\label{sec:exs}
In this paper, the following experimental setup was used. The primary sensor is a preproduction unit of Sonar~3D-15 from Waterlinked~\cite{waterlinked}. 
The Sonar 3D-15 is an advanced 3D sonar with approximately a $90^\circ$ by $40^\circ$ field of view in navigation mode, the only mode tested and available at the time of this reporting. 
The navigation mode provides a range of up to 15 meters with an update frequency of \SI{5}{Hz}. The sensor was rigidly mounted on a BlueRov2 platform~\cite{BlueROV2}. 
The BlueRov2 is equipped with a Raspberry PI camera, an IMU and a pressure sensor measuring the current depth, thus providing globally consistent measurements along the Z-axis. 
Furthermore, a GoPro Hero Black (10 and 13) with an internal IMU is mounted next to the sonar, facing forward. 
Illumination is provided by four 1500-lumen Subsea Light~\cite{lumen} and a 20000-lumen Keldan light~\cite{keldan}. Fig.~\ref{fig:beauty} shows the experimental setup deployed inside a cavern in Florida.   

\begin{figure*}[h]
    \centering
    \includegraphics[width=2.0\columnwidth]{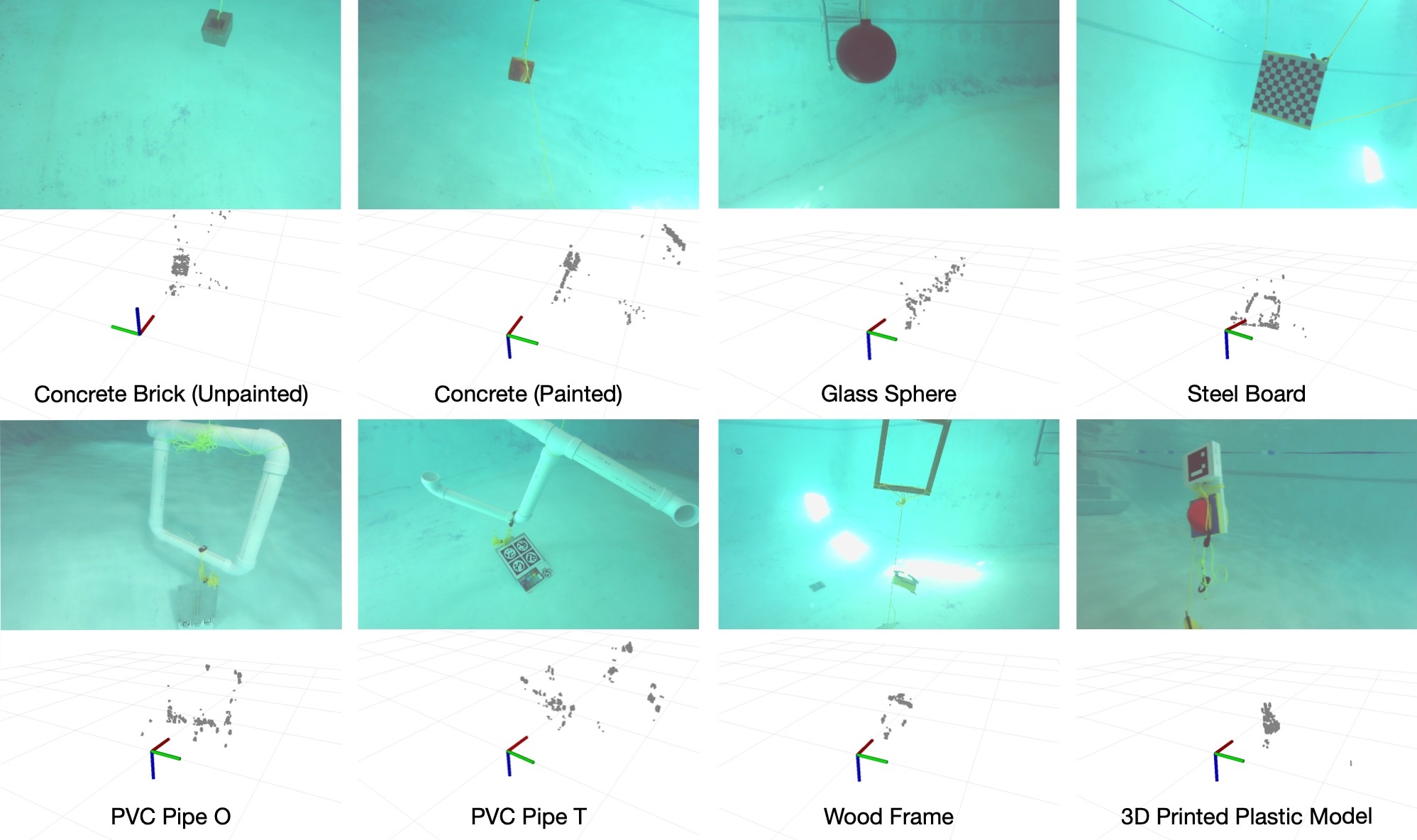}
    \caption{The sonar response on different materials commonly found in underwater environments and used for calibration.}
    \label{fig:Materials}
\end{figure*}

For this study, four datasets were utilized, each presenting different characteristics:
\begin{itemize}
    \item Pier Pilling: wood piles in shallow water, collected with the BlueROV2 teleoperated.
    \item Container: metal submerged tank with opening.
    \item Peacock Springs Cave: long cave structure, collected with the BlueROV2 with a mounted GoPro13, while handheld by cave divers. The diver entered and exited from the same area, forming a trajectory with a start-end loop.
    \item Ginnie Springs cavern: a cavern with a larger opening, variability in depth, and the presence of narrower areas. The divers carried the BlueROV2 with 3D sonar and the GoPro10, and formed loops. The dataset is composed of two days, day I with the powerful  Keldan light on, and day II with the less powerful lights mounted on the BlueROV2.
\end{itemize}
%
\subsection{Sonar Characterization and Material Response}
During the experiments the sonar operated in low (\SI{1.2}{MHz}) frequency, producing a beam of \ang{0.6} horizontal and \ang{2.4} vertical angle. The returned range measurements have \SI{4}{mm} resolution. Physically, the Sonar 3D-15 measures \SI{0.08}{m} height by \SI{0.12}{m} width by \SI{0.04}{m} depth and weights \SI{390}{g} submerged.   

To select a suitable material for sonar–camera calibration and assess the performance of the 3D sonar with different surfaces, experiments were conducted to evaluate 3D sonar responses across different materials. 
Selected materials included those frequently encountered in underwater environments, such as bricks, concrete, PVC pipes, and wood.
Additionally, easily available items typically used for calibration were included, such as a glass sphere, popular for 2D sonar calibration, and a metal checkerboard with 3D-printed tag for camera calibration.
The camera image and corresponding sonar point-cloud of the test materials are shown in Fig.~\ref{fig:Materials}.
The experiment was conducted in an indoor swimming pool with smooth walls, which introduced multipath effects in the collected data. 
The test materials were suspended in the middle of the water to minimize the influence of the pool walls.
To further mitigate their impact on the material experiment results, the sensor range was limited to $r = 2\text{m}$ for visualization in Fig.~\ref{fig:Materials}.

The results demonstrated that the responses varied with both material type and surface properties at different viewing angles.
While the glass sphere and steel calibration board provide strong reflections for the 2D sonar, they are not well-suited for 3D sonar calibration due to multipath effects.
Regarding surface properties, smooth objects such as PVC pipe produced noisier detections due to reflections, while rougher materials such as concrete yielded less noisy measurements.
Both spray painted and unpainted concrete produced relatively strong signals.
%
\subsection{Calibrating the extrinsic parameters}
\label{subsec:calib}

\begin{figure}[h]
    \centering
    \subfigure[]{%
        \includegraphics[height=0.12\textheight, width=0.233\textwidth, trim={0.0in, 0.0in, 2.0in, 0.0in},clip]{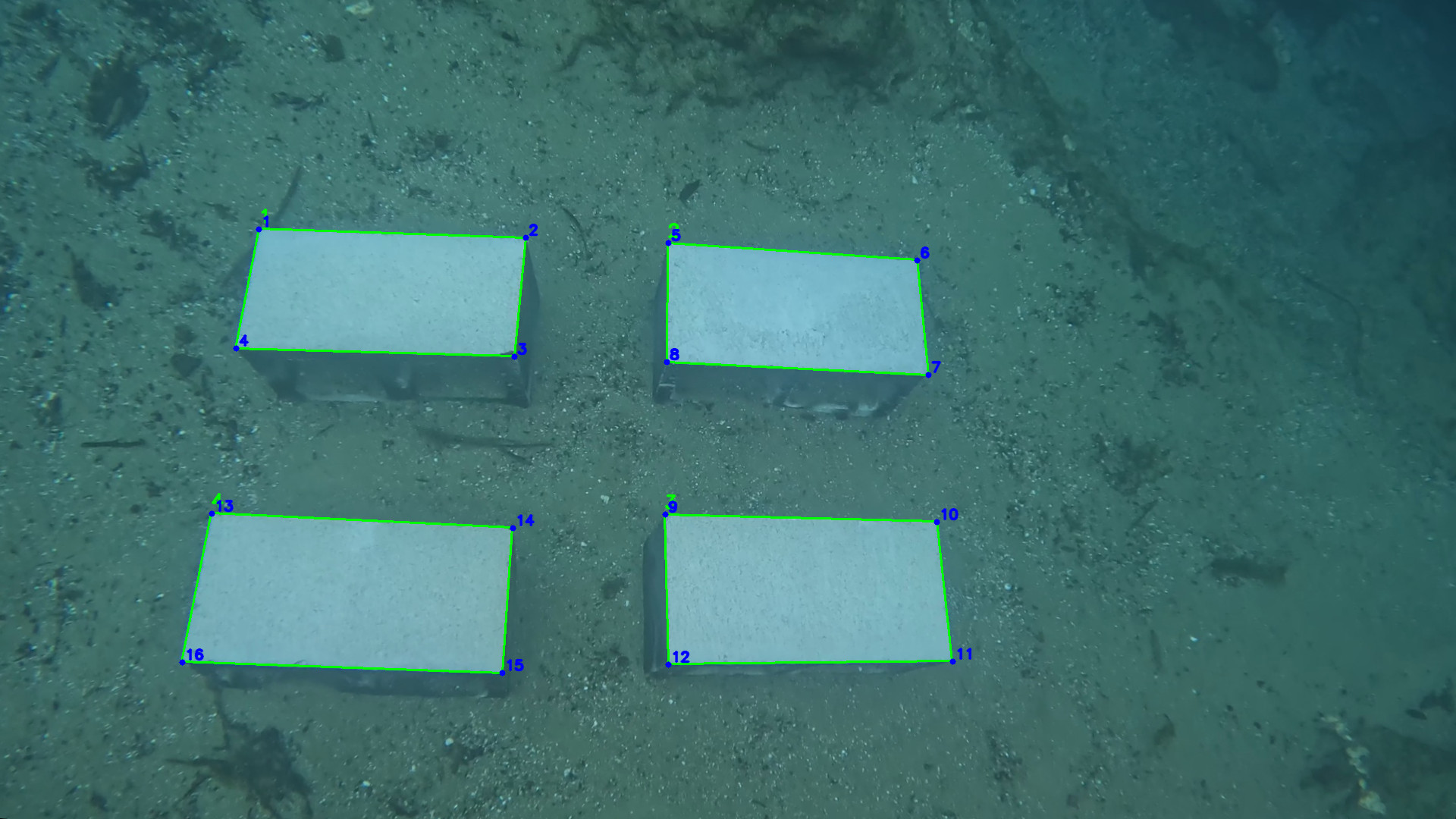}%
        \label{fig:calibration:a}%
    }
    \subfigure[]{%
        \includegraphics[height=0.12\textheight, width=0.233\textwidth, trim={0.0in, 2.5in, 0.0in, 0.0in},clip]{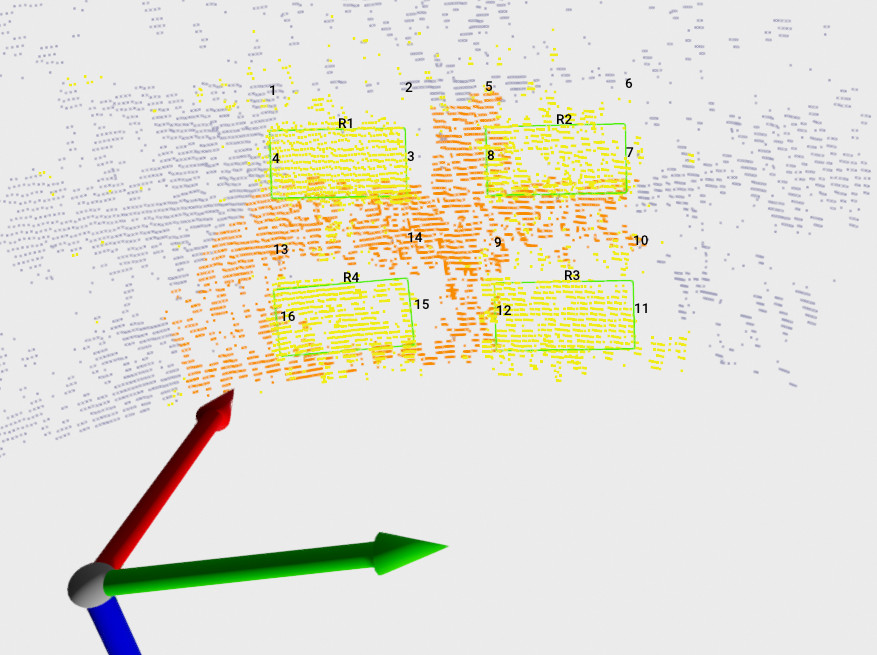}%
        \label{fig:calibration:b}%
    }
    \caption{Calibration setup. (a) Camera image of the four cinder blocks showing the top-surface borders marked in green and detected corners in blue. (b) Pointcloud from the Sonar 3D-15 displaying detected cinder blocks in yellow with borders highlighted in green.}
    \label{fig:Calibration}
\end{figure}

Based on the material experiments, cinder (concrete) blocks (Fig. \ref{fig:calibration:a} were selected due to their rough surface texture, well-defined geometric structure, and convenient availability. 
Four cinder blocks were placed at that the bottom of the water body in a particular arrangement, such that the 16 corners of top surfaces form a distinctive pattern. 
Non-top surfaces were spray painted to suppress spurious edges and aid corner detection in the camera images.
A rectangular mask is used to identify the corners of the four cinder blocks, which are then labeled correspondingly in the visual images. 
The top surface of the cinder blocks are extracted from sonar point-clouds by identifying the elevated planar regions.
As shown in Fig. \ref{fig:calibration:b}, points on these elevated surfaces were subsequently denoised and clustered into four separate 2D surfaces.

\begin{figure}[h]
    \centering
    \includegraphics[width=1.0\columnwidth, trim={0.0in, 2.0in, 0.0in, 2.0in},clip]{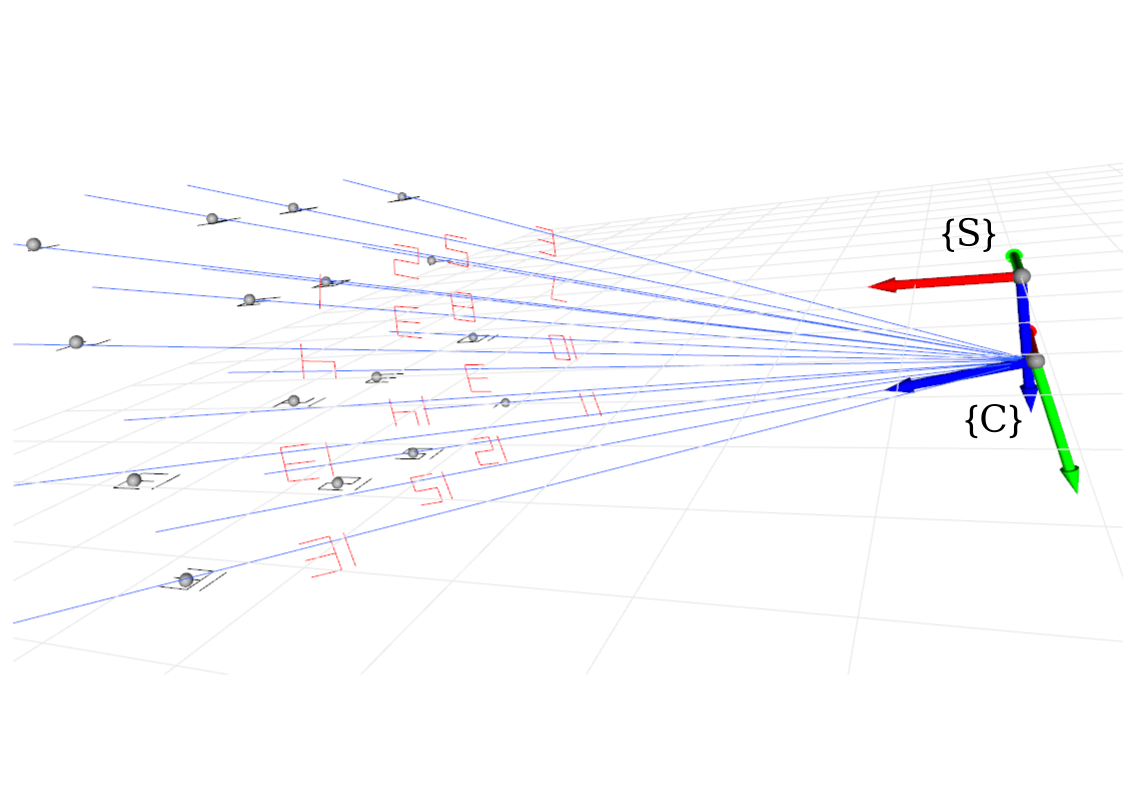}
    \caption{Rays from the camera projected through 16 cinder block corners in the 2D image plane to their corresponding 3D points. Sonar 3D corner points are annotated in black and visual points in red. The camera frame $\{C\}$ and sonar frame $\{S\}$ are also illustrated in the figure using the calibrated extrinsic transformation.}
    \label{fig:PnP}
\end{figure}

After detecting the corner points of the cinder blocks in both the visual images and sonar pointcloud, correspondences between them are established to form a 3D–2D association.
Define a sonar 3D point as $\mathbf{X} = [x, y, z]^{\top}$, and a visual point as $\mathbf{p} = [u,v]^{\top}$. The projection model, with $s$ a scalar scale factor, is given by:
\begin{equation}
s \begin{bmatrix}  \mathbf{p} \\ 1
\end{bmatrix}
= \mathbf{K} \cdot [\mathbf{R}| \mathbf{t}]
\begin{bmatrix} 
\mathbf{X} \\ 1 
\end{bmatrix}
\end{equation}
The intrinsic matrix $\mathbf{K}$ is obtained through standard camera calibration. 
The undistorted visual point $\mathbf{p}'$ is defined as $[\mathbf{p}' ,1]^{\top} = \mathbf{K}^{-1} [ \mathbf{p}, 1]^{\top}$. The camera-to-sonar extrinsic $(\mathbf{R,t})$ calibration problem can then be formulated as a Perspective-n-Point (PnP) problem using 16 brick corner correspondances.
We solve this problem using EPnP\cite{lepetit2009ep} algorithm.
Fig. \ref{fig:PnP} illustrates the alignment of camera rays passing through 16 brick corners in the 2D image plane and their corresponding 3D points, along with the estimated camera–sonar extrinsic.


\section{Method Description}
\label{sec:app}

\subsection{3D Sonar Scan Registration}

Since the pointcloud collected by a compact 3D sonar shares many characteristics with solid-state 3D LiDAR, 3D sonar scan registration is performed using the state-of-the-art LiDAR scan registration algorithm, KISS-ICP \cite{vizzo2023kiss}.
KISS-ICP is a point-to-point ICP algorithm designed to provide a general solution across different LiDAR platforms and application environments. 
In this paper, we extend the use of KISS-ICP from various LiDAR configurations to 3D sonar (termed Sonar-odom) and evaluate its performance in a series of underwater scenarios.
Since the sonar pointcloud is much sparser than the LiDAR pointcloud, we skipped the down-sampling of the source pointcloud and only perform down-sampling on the target pointcloud.

KISS-ICP adopts a constant velocity model for generality.
At a current timestamp $n$, the initial guess $\mathbf{T}_{\text{init}, n}$ transformation matrix for ICP scan registration is calculated based on the pose estimates $\mathbf{T}_{n-1}$, $\mathbf{T}_{n-2}$ from previous timestamps:
\begin{equation}
    \mathbf{T}_{\text{init}, n} = \mathbf{T}_{n-1} \cdot (\mathbf{T}_{n-2}^{-1} \cdot \mathbf{T}_{n-1})
\end{equation}

However, ICP is sensitive to the selection of initial guess\cite{rusinkiewicz2001efficient}.
Compared to regular 3D LiDARs, which have a wider field of view and higher frequency, the Sonar 3D-15 has a maximum range of only 15 meters with \SI{5}{Hz} update rate.
Furthermore, the complex hydrodynamic conditions in real underwater environments introduce additional uncertainties in the scan registration of the 3D sonar data.
As the robot platform utilized is equipped with both a 3D sonar and a GoPro camera (Fig.~\ref{fig:beauty}, left), we provided the 3D sonar a more robust initial prior using visual-inertial odometry (VIO) from SVIn \cite{RahmanIJRR2022}, which we call Sonar-odom-SVIn-Initial.
The VIO estimation in SVIn is performed at a higher frame rate (\SI{30}{Hz} camera rate). 
Therefore, we adopt the SVIn odometry from the closest timestamp, $\mathbf{T}_{\text{VIO}, n'}$ (with $n' \approx n$), as the initial guess for 3D sonar ICP scan registration. 
The odometry pose is transformed into the sonar frame, $s$, using the extrinsic parameters $\Delta \mathbf{T}$ obtained in Section \ref{sec:exs}:
\begin{equation}
    {}_s\mathbf{T}_{\text{VIO}, n'} = \Delta \mathbf{T} \cdot \mathbf{T}_{\text{VIO}, n'} \cdot \Delta \mathbf{T}^{-1}
\end{equation}
The initial guess for the 3D sonar scan registration is then defined as:
\begin{equation}
    \mathbf{T}_{\text{init}, n} = \mathbf{T}_{n-1} \cdot ({}_s\mathbf{T}_{\text{VIO}, n''}^{-1} \cdot {}_s\mathbf{T}_{\text{VIO}, n'})
\end{equation}
where $n'' \approx n-1$ denotes the VIO timestamp closest to $n-1$.

\subsection{3D Sonar Loop-Closure and Pose Graph Optimization}
Due to the relative limited field of view and low update frequency of the sonar, the odometry-only trajectory shows significant drift, especially over longer durations (Table~\ref{tab:ate}).
An additional strategy to mitigate drift is loop-closure detection and pose graph optimization, which we call \emph{Sonar PGO}.
Similar to LiDAR SLAM\cite{shan2020lio}, we detect loop-closures using a Euclidean distance-based method.
Keyframes, denoted as $\mathcal{F} = \{\mathbf{F}_i\}$, are selected to ensure a stable and efficient performance.
If the current source keyframe $\mathbf{F}_s$ is geometrically close enough to, and temporally sufficiently separated from a target keyframe, $\mathbf{F}_t$, a potential loop-closure is detected.
We then perform scan-to-map matching using the target sub-map, $\mathcal{M}_t = \{\mathbf{F}_{t-w}, \dots, \mathbf{F}_{t+w}\}$, with a window size of $w = 10$.

In pose graph optimization, the sonar pose at each keyframe $\mathbf{F}_i$ is treated as a state $\mathbf{X}_i$.
Sonar odometry from scan registration is modeled as a sequential odometry measurement $\mathbf{Z}^{o}_{i-1, i}$, while the scan-to-map loop-closure is modeled as a loop-closure measurement $\mathbf{Z}^{l}_{i, j}$.
These measurements are added to the graph as factors $f(\mathbf{X}_{i-1}, \mathbf{X}_i; \mathbf{Z}^{o}_{i-1, i})$ and $f(\mathbf{X}_{i}, \mathbf{X}_j; \mathbf{Z}^{l}_{i, j})$.
The overall factor graph at a given time $n$, with $m$ detected loop-closures, can be expressed as: 
\begin{equation}
f(\mathcal{X}_n; \mathcal{Z}_n) = f(\mathbf{X}_0) 
\prod^n f(\mathbf{X}_{i-1}, \mathbf{X}_{i};\mathbf{Z}^{o}_{i-1, i})
\prod^m f(\mathbf{X}_i, \mathbf{X}_j; \mathbf{Z}^{l}_{i, j})
\end{equation}
with $\mathcal{X}_n = \{\mathbf{X}_i\}$ denoting the set of all states from time $0$ to $n$, and $\mathcal{Z}_n = \mathcal{Z}_n^o \cup  \mathcal{Z}_n^l$ denoting the set containing both odometry measurements $\mathcal{Z}_n^o = \{\mathbf{Z}_{i-1, i}^o\}$ and loop-closure measurements $\mathcal{Z}_n^l = \{\mathbf{Z}_n^l\}$.

Note we refer to the use of SVIn as an initial guess for sonar registration and loop-closure as the \textbf{Fusion approach}. While SVIn originally denotes Sonar Visual-Inertial Navigation, in this work we utilize only its visual-inertial odometry (VIO) component~\cite{leutenegger2015keyframe}, since the sonar sensor used in SVIn2~\cite{RahmanIJRR2022} was a pipe profiling sonar, which differs from our setup.

In this section, we present the performance of Sonar SLAM, SVIn VIO and our fusion approach from localization and mapping perspectives, evaluated both qualitatively and quantitatively across various datasets.
We select three representative datasets for comparison: the Ginnie Springs cave system, the Peacock Springs cave system, and the container dataset.
The Ginnie Springs dataset contains floating particulates in the images, bringing challenges for the state-of-art underwater VIO framework.
The Peacock Springs dataset is a long-duration dataset that demonstrates the ability of our system to handle extended localization and mapping missions.
The container dataset represents a challenging environment that includes a steel container. As discussed in Sec.~\ref{sec:exp}, the steel would cause multi-path effect for the 3D sonar, making localization and mapping using sonar pointcloud even more challenging.

%
\subsection{Sonar SLAM and Fusion Approach}
We evaluate the performance of Sonar SLAM (referred to as Sonar PGO) and our visual-sonar fusion approach (referred to as Fusion PGO), comparing them to the state-of-art underwater VIO, SVIn2 (referred to as SVIn PGO), on datasets collected in the Ginnie Springs cave system.

\begin{figure}[t]
    \centering
    \includegraphics[width=1.0\columnwidth, trim={0.0in, 1.6in, 0.0in, 2.36in},clip]{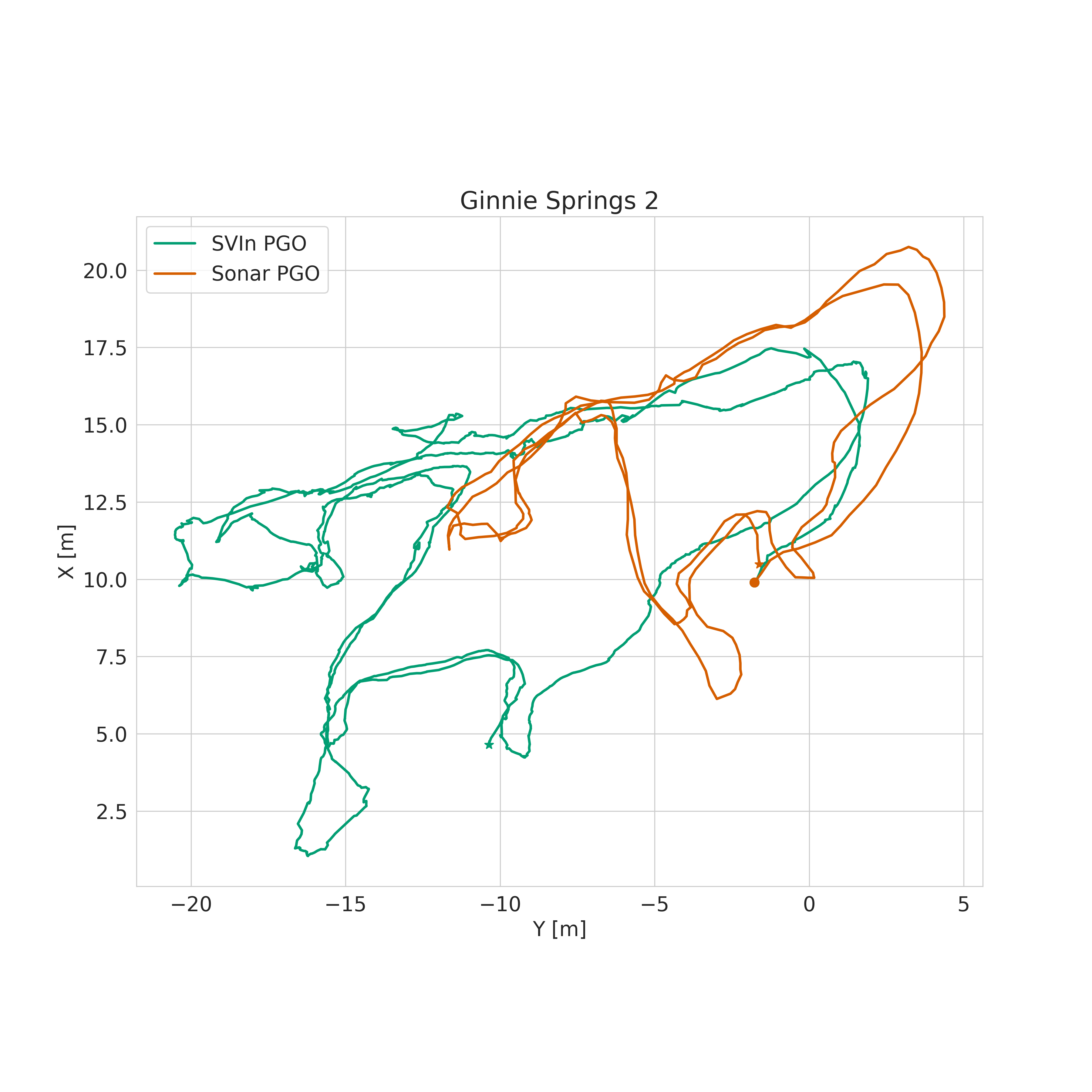}
    \caption{Trajectories comparison of sonar-only and SVIn on the Ginnie ballroom I dataset.}
    \label{fig:sonar_vs_svin_traj_2}
\end{figure}

\begin{figure}[h]
    \centering
    \subfigure[]{%
        \includegraphics[height=0.14\textheight, width=0.233\textwidth, trim={0.0in, 0.0in, 2.0in, 0.0in},clip]{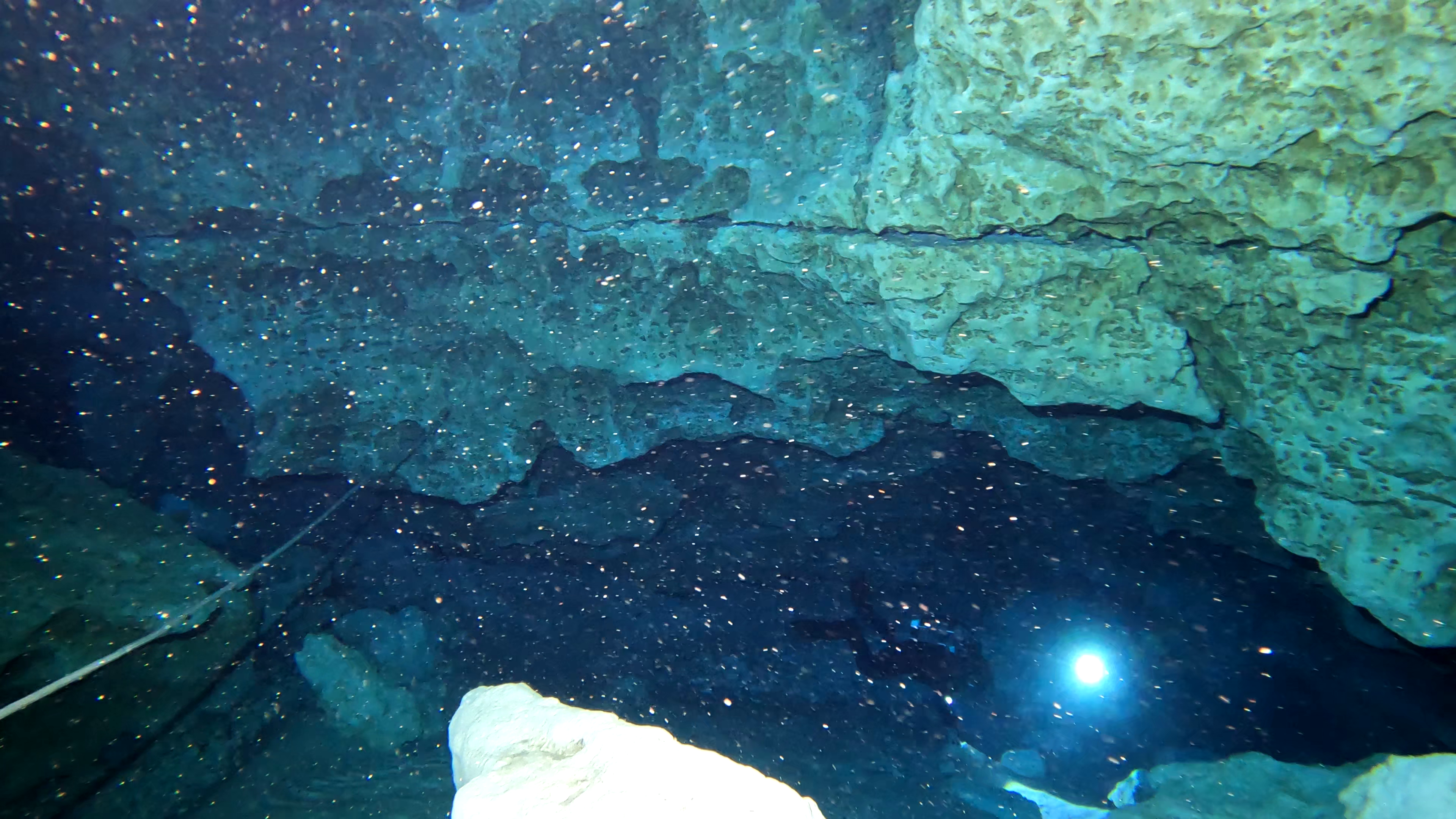}%
        \label{fig:particle:a}%
    }
    \subfigure[]{%
        {\includegraphics[height=0.14\textheight, width=0.233\textwidth, trim={1.4in, 0.0in, 1.0in, 0.3in},clip]{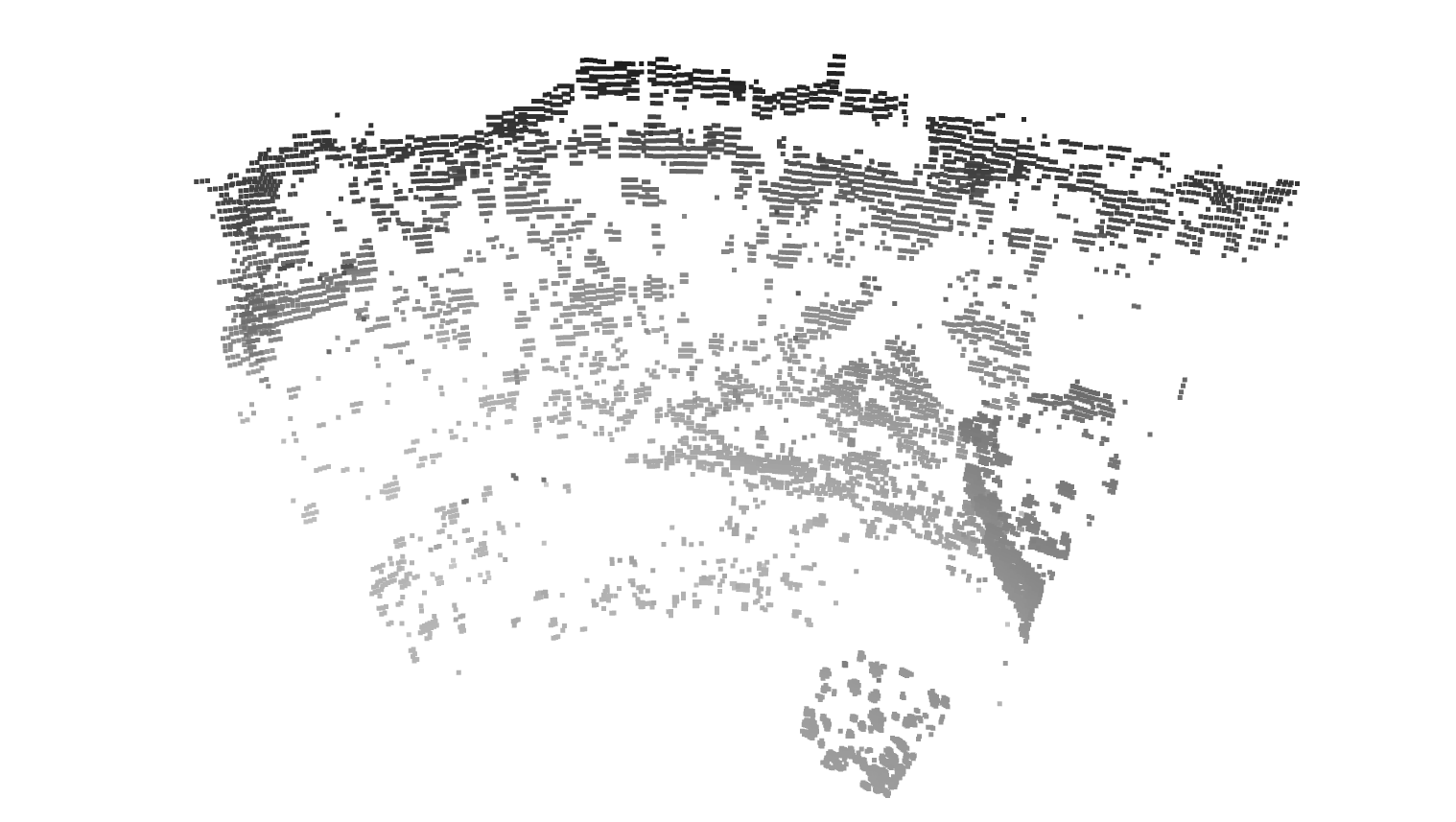}}%
        \label{fig:particle:b}%
    }
    \caption{GoPro camera image (a) and sonar pointcloud (b) captured at nearly the same timestamp. The camera image is filled with floating particulates, whereas the sonar pointcloud is unaffected by them.}
    \label{fig:particulates}
\end{figure}

\begin{figure}[h]
    \centering
    \includegraphics[width=1.0\columnwidth, trim={0.0in, 0.6in, 0.0in, 1.42in},clip]{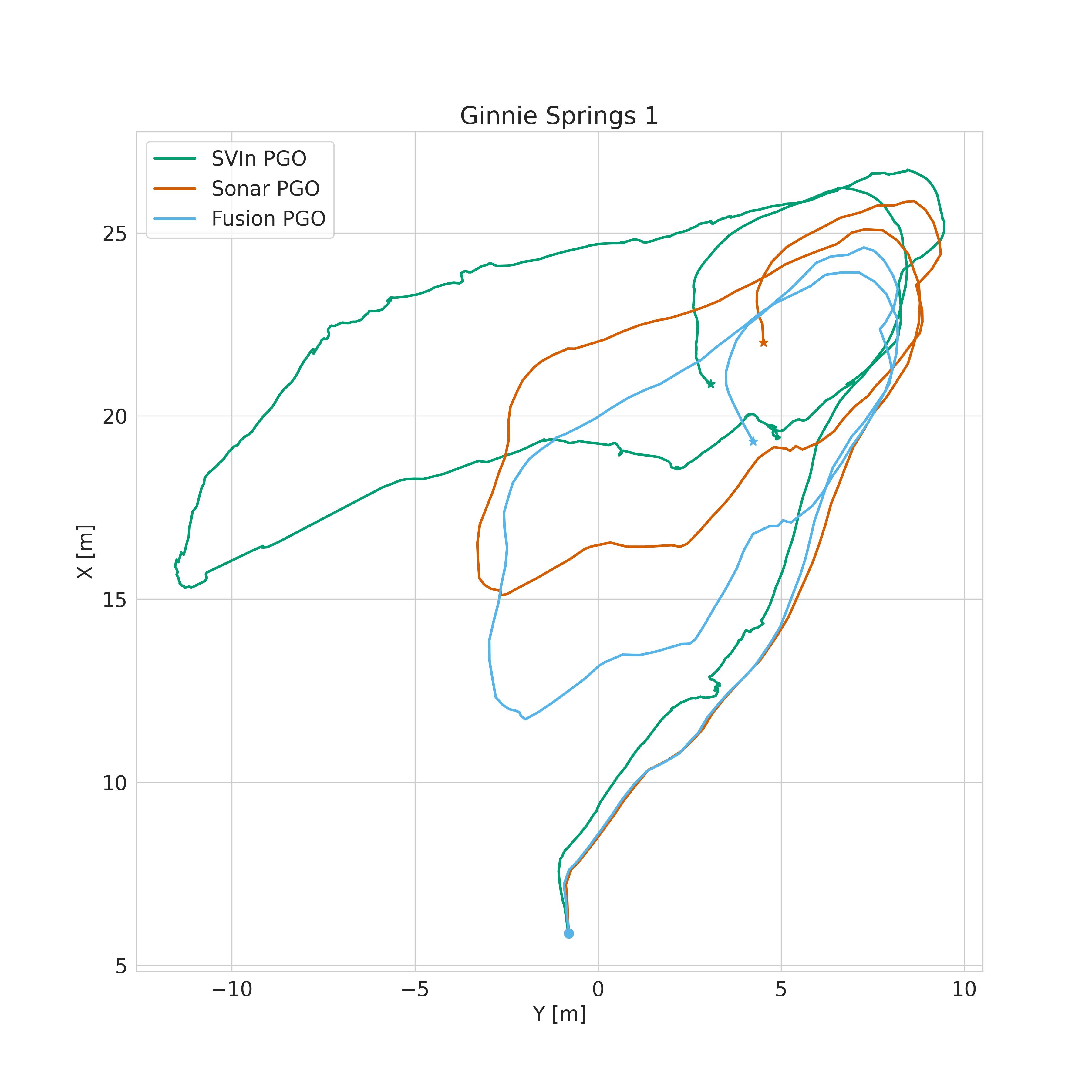}
    \caption{Trajectories comparison of our Fusion approach, SVIn and Sonar-only on the Ginnie ballroom II dataset.}
    \label{fig:sonar-svin_traj_1}
\end{figure}

\section{Experimental Validation}
\label{sec:exp}

Fig.~\ref{fig:sonar_vs_svin_traj_2} presents the comparison between Sonar PGO and SVIn PGO on a start-end loop trajectory, referred to here as Ginnie ballroom I dataset.
All trajectories are aligned using the keyframe poses from first five timestamps of the Sonar PGO with EVO\footnote{EVO: \url{https://github.com/MichaelGrupp/evo}}, and the starting and ending points are marked by a circle and a star, respectively.
Due to the presence of floating particulates (Fig. \ref{fig:particulates}), which are detected and misinterpreted as static features, SVIn exhibits a larger start–end re-visitation drift ($10.34~\mathrm{m}$) compared to Sonar PGO ($3.31~\mathrm{m}$).
Though raw sonar scan matching suffers from the drift caused by the low sensor update frequency and the constant-velocity motion model used by Kiss-ICP, the longer sonar range (15 meters) ensures that enough loop closures are detected, thereby constraining the overall drift and achieving a better start-end re-visitation drift error.
%

We evaluate the performance of our fusion approach on the datasets collected in Ginnie Ballroom II, 
and compared it against two configurations, Sonar-only SLAM and SVIn.
Fig.~\ref{fig:sonar-svin_traj_1} shows the trajectories using various configurations.
Similarly, all trajectories are aligned using the first five timestamps of Sonar PGO.
Incorporating the SVIn prior into Sonar-PGO reduces overall drift compared to both methods on the Ginnie Ballroom II dataset. 
THe vision-based method drifts when there are floating particulates, whereas the sonar remains unaffected by such occlusions and provides reliable longer range measurements. However, sonar-only approaches suffer from registration errors due to low measurement frequency. This drawback is mitigated by the higher-frequency odometry from SVIn, which supplies robust motion priors for scan alignment.
By combining both camera and sonar data, the system exhibits lower drift compared to trajectories using only the camera or the sonar.
%

%

\subsection{Dense Pointcloud Reconstruction}
\subsubsection{Reconstruction of the Cave System}
\begin{figure}[h!]
    \centering
    
    \includegraphics[width=1.0\columnwidth, height=5cm, trim={0.0in, 0.0in, 0.25in, 0.0in},clip]{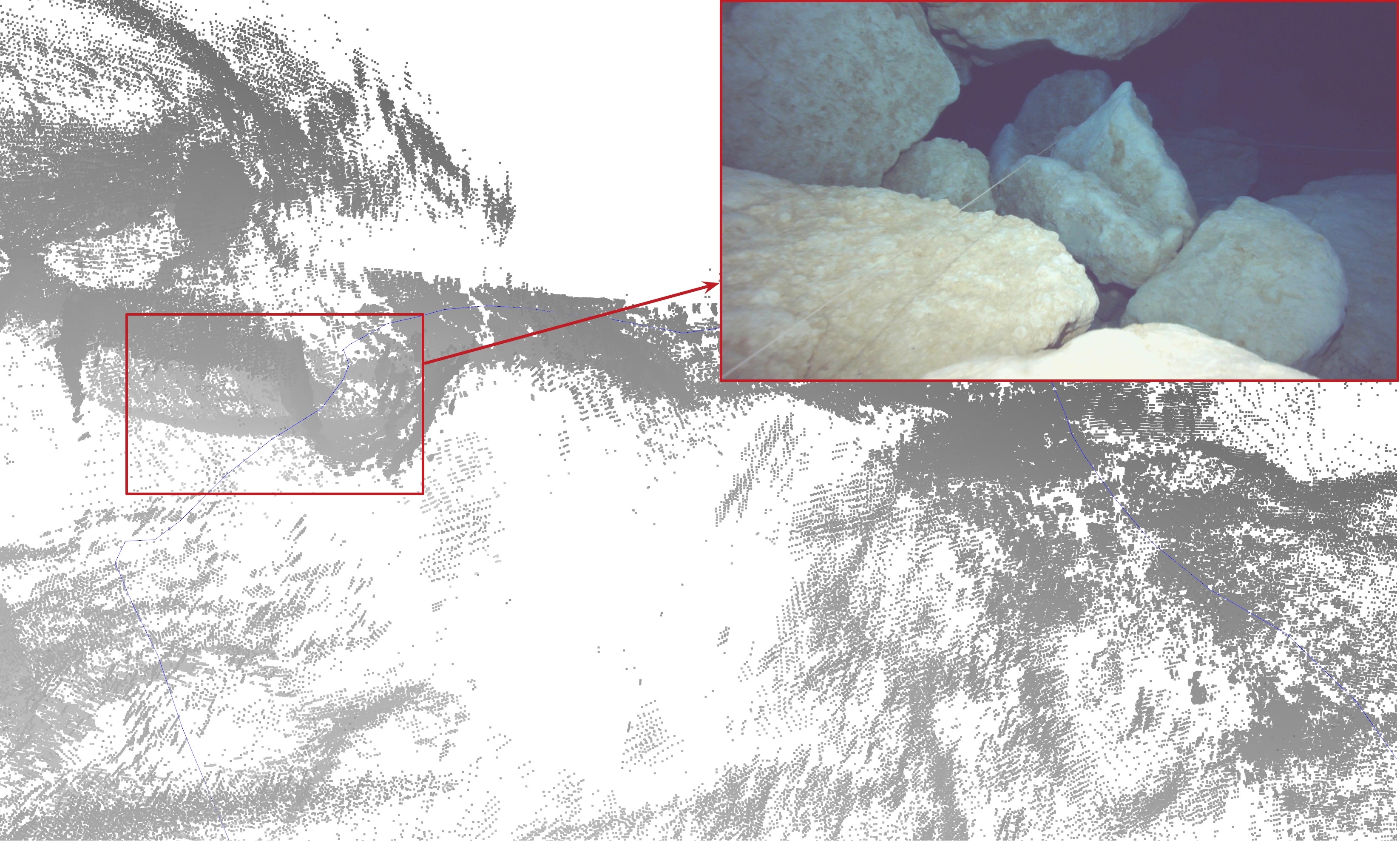}
    \caption{Detailed pointcloud reconstruction of the cave interior from the Ginnie Ballroom I dataset, using a limestone structure as a reference.}
    \label{fig:gennie_cave_pc}
\end{figure}

\begin{figure}[t!]
    \centering
    \includegraphics[width=1.0\columnwidth, height=12cm,
                     trim={0.0in, 0.0in, 0.25in, 0.0in},clip]{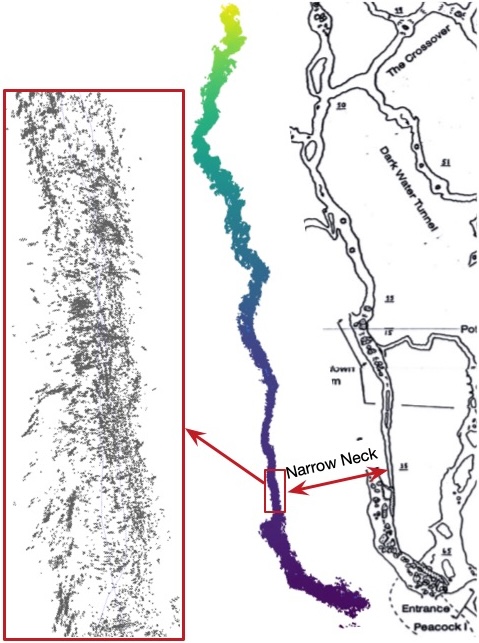}

    \vspace{1mm}
    \begin{tabular*}{\columnwidth}{@{\extracolsep{\fill}}ccccccccccccccccccc}
       & & & & (a)& & & & & & & (b)& & &  &  & (c) &  &  \\
    \end{tabular*}
    \caption{Results of Fusion approach on a 700\,m trajectory inside Peacock Springs cave system: 
    (a) detailed pointcloud of the narrow neck, (b) pointcloud map of the explored cave system, and (c) the corresponding Cave Atlas map.}
    \label{fig:peacock_caves_map}
\end{figure}

\begin{figure}[t!]
    \centering
    \subfigure[]{%
        \includegraphics[width=0.49\columnwidth, trim={0.0in, 0.0in, 0.0in, 0.0in},clip]{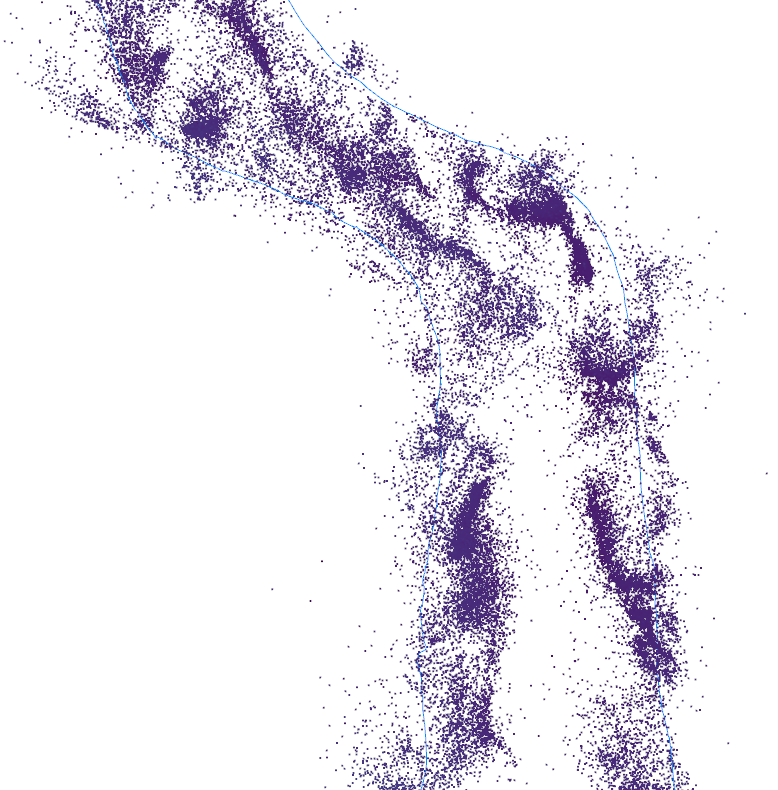}%
        \label{fig:peacock_caves_map2:a}%
    }
    \subfigure[]{%
        \includegraphics[width=0.49\columnwidth, trim={0.0in, 0.0in, 0.0in, 0.0in},clip]{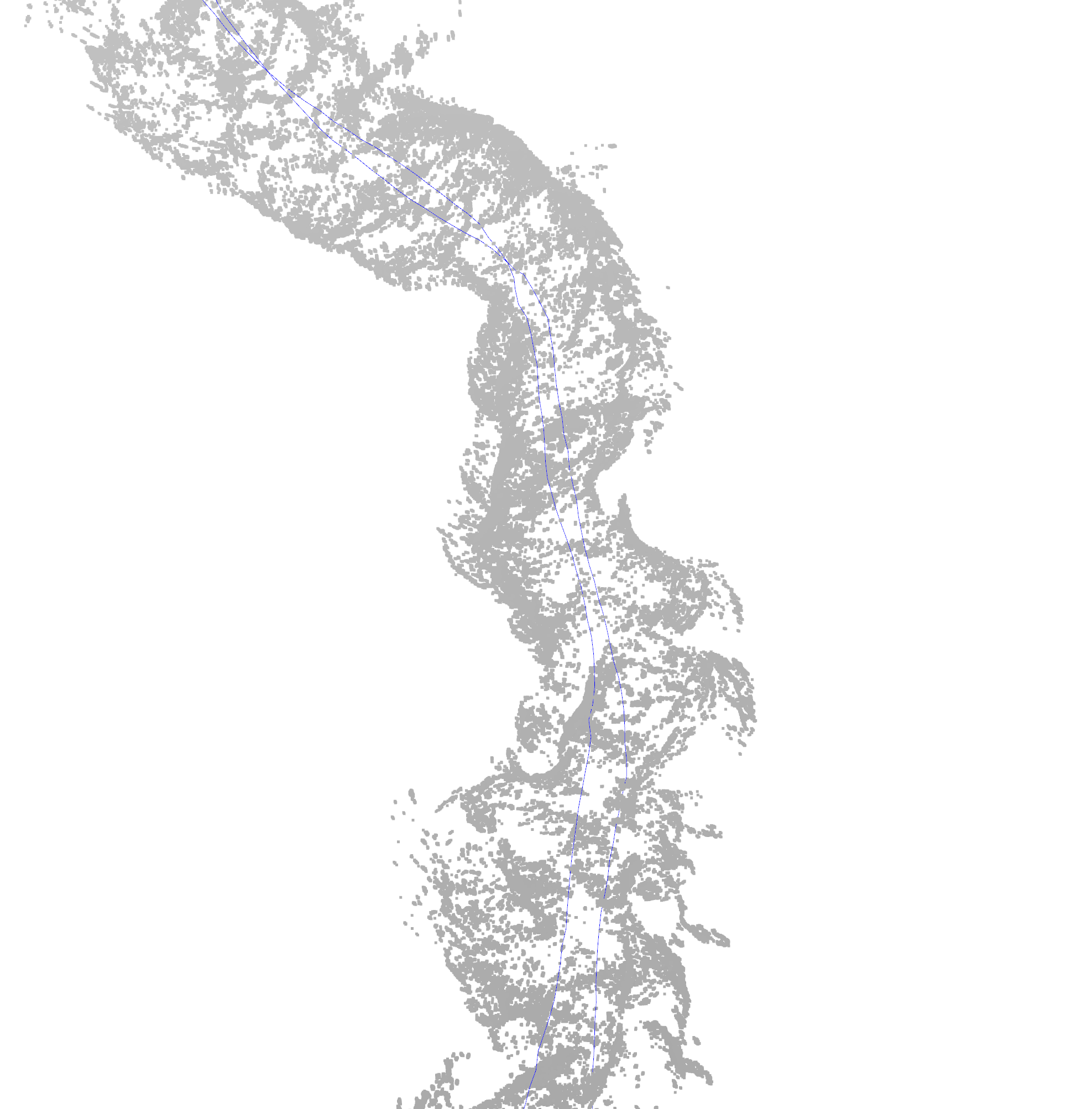}%
        \label{fig:peacock_caves_map2:b}%
    }
    \caption{Comparison of detailed pointcloud reconstructions of the same area in Peacock Caves: (a) SVIn PGO and (b) Fusion PGO approach.}
    \label{fig:peacock_caves_map2}
\end{figure}

We also demonstrate the dense reconstruction results of our fusion approach in the challenging environments  of the Ginnie Springs cavern (Fig. \ref{fig:gennie_cave_pc}) and the Peacock Springs (Fig. \ref{fig:peacock_caves_map} and \ref{fig:peacock_caves_map2}).

The Sonar 3D-15 enables consistent dense reconstruction, as shown in a zoomed-in view of the interior from the Ginnie Ballroom.
As mentioned previously, this dataset contains floating particulates that interfere with the performance of SVIn.
However, in the sonar reconstruction results, some basic topographic features of the cave environment are perceptible, demonstrating the effectiveness of our fusion and loop closure detection strategy.
In Fig. \ref{fig:gennie_cave_pc}, the limestone structure and a nearby narrow passage can be roughly identified in the dense point cloud, corresponding well to the collected image.


The Peacock Spring dataset presents confined, tunnel-like settings, posing significant difficulties for conventional 3D sensing modalities.
Although SVIn achieves a relatively low drift rate, as shown in Tab.~\ref{tab:ate}, failures in loop closure due to sparse visual features and viewpoint differences, as illustrated in Fig.~\ref{fig:peacock_caves_map2:a}, prevent successful loop closure.
In contrast, the fusion approach is largely unaffected by these limitations. As shown in Fig.~\ref{fig:peacock_caves_map}, it effectively reconstructs even the narrowest part of the dense structural layout of the confined cave environment.  
We generate the dense pointcloud from sonar pointclouds associated with the keyframes in the pose graph optimization.
Compared to the existing Cave Atlas maps, which are manually surveyed and contain only the topographic relationships of the caves, the reconstructed pointcloud map provides finer structural detail, improved scale accuracy, and a full 3D representation.

\begin{table}[htp]
\caption{\textbf{Re-visitation Error [m]} between the start and end points of the Peacock Springs cave system trajectory under different methods and configurations.} 
\renewcommand{\arraystretch}{1.2}
\label{tab:ate}
\begin{center}
\setlength{\tabcolsep}{3pt}
\begin{tabular}{ccccc}
\Xhline{2\arrayrulewidth}
{\textbf{\footnotesize SVIn PGO}} & \textbf{\footnotesize Sonar Odom} & {\textbf{\footnotesize Sonar PGO}} & \textbf{\footnotesize Fusion Odom} & {\textbf{\footnotesize Fusion PGO}}\\ 
\Xhline{2\arrayrulewidth}
 17.02 & 250.34 & 21.51 & 238.71 & 6.76\\ 
\Xhline{2\arrayrulewidth}
\end{tabular}
\end{center}
\end{table}

The re-visitation error for the start–end loop of around a 700-meter trajectory collected in one of the Peacock Springs cave systems is summarized in Tab.~\ref{tab:ate}.
The error is calculated as the Euclidean distance between the starting and ending points of the trajectory.
The cave entrance has a diameter of approximately $10$ meters; therefore, even if the starting and ending points do not coincide, the re-visitation error should be no greater than $10$ meters. Our fusion approach achieves a re-visitation error of 6.76 meters, which is the closest to this theoretical bound.
%

\subsubsection{Operation in Acoustically Challenging Environments}

\begin{figure}[t!]
    \centering
    \subfigure[]{%
        \includegraphics[width=0.49\linewidth, trim={0.0in, 0.0in, 0.0in, 0.0in},clip]{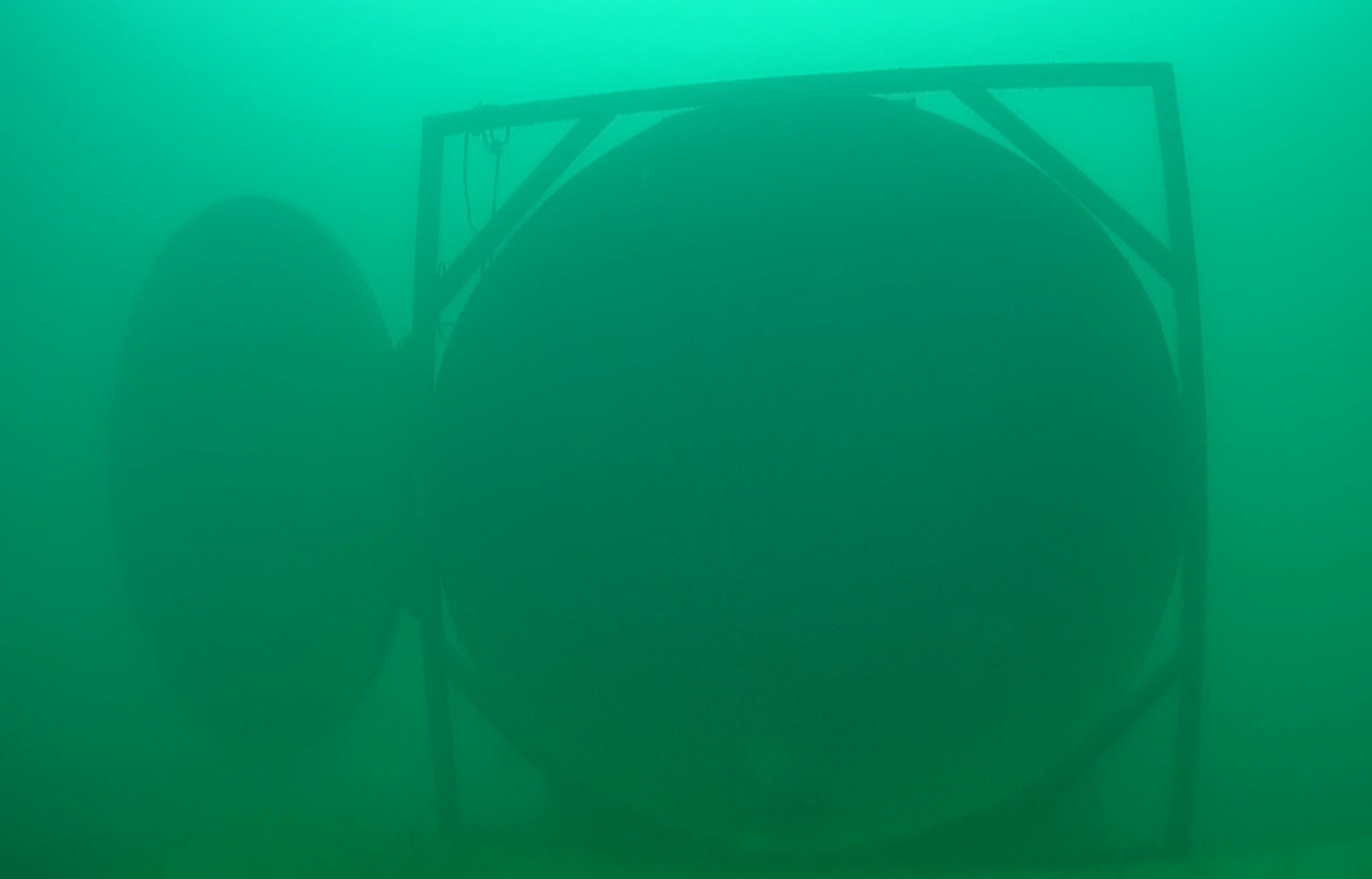}%
        \label{fig:pipe-container:a}%
    }
    \hfill
    \subfigure[]{%
        \includegraphics[width=0.49\linewidth, trim={6.0in, 6.0in, 11.0in, 0.0in},clip]{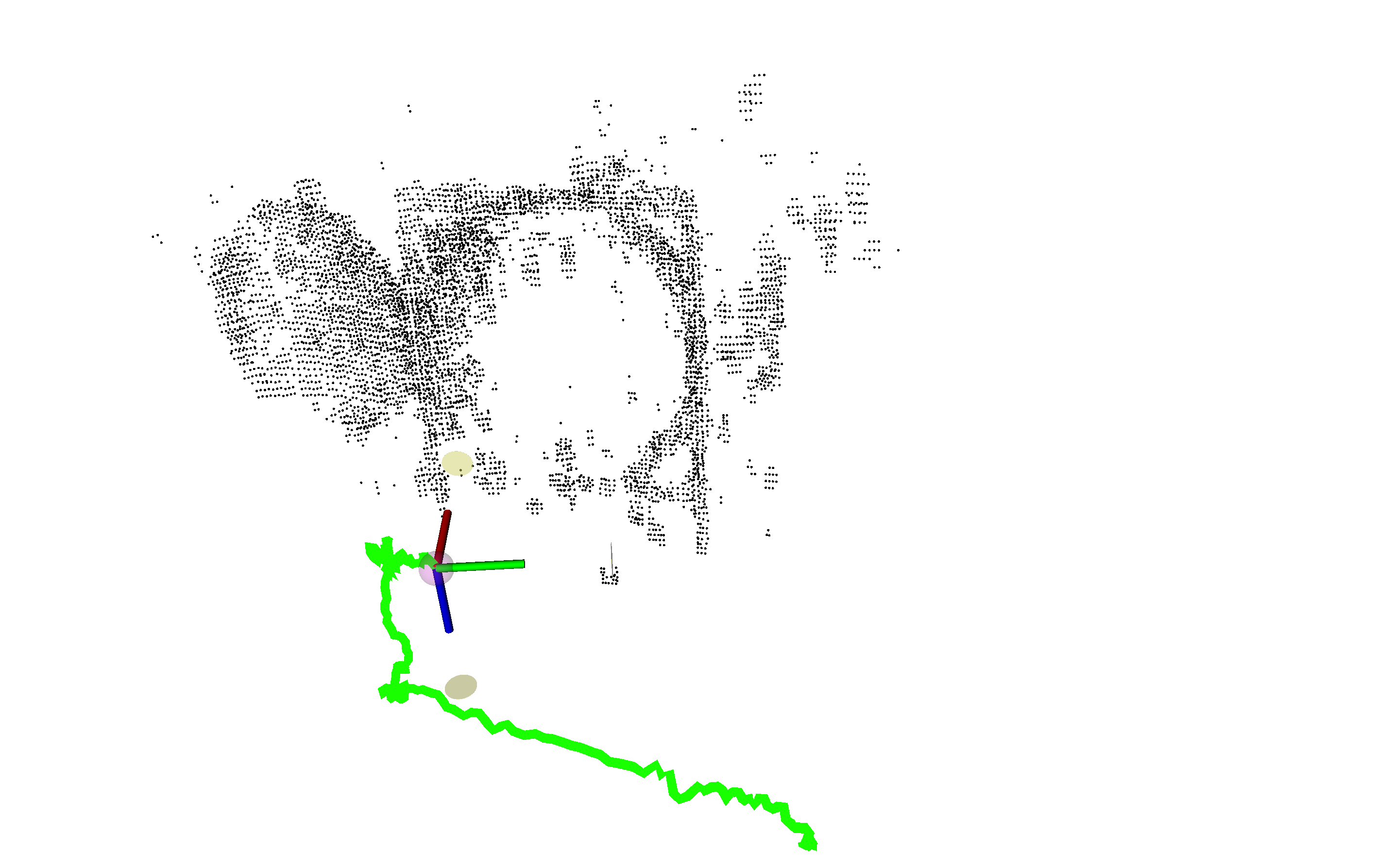}%
        \label{fig:pipe-container:b}%
    }
    
    \vspace{0.7mm} 

    \subfigure[]{%
        \includegraphics[width=\linewidth,height=0.97\linewidth, trim={0.5in, 1.0in, 4.0in, 0.0in},clip, keepaspectratio]{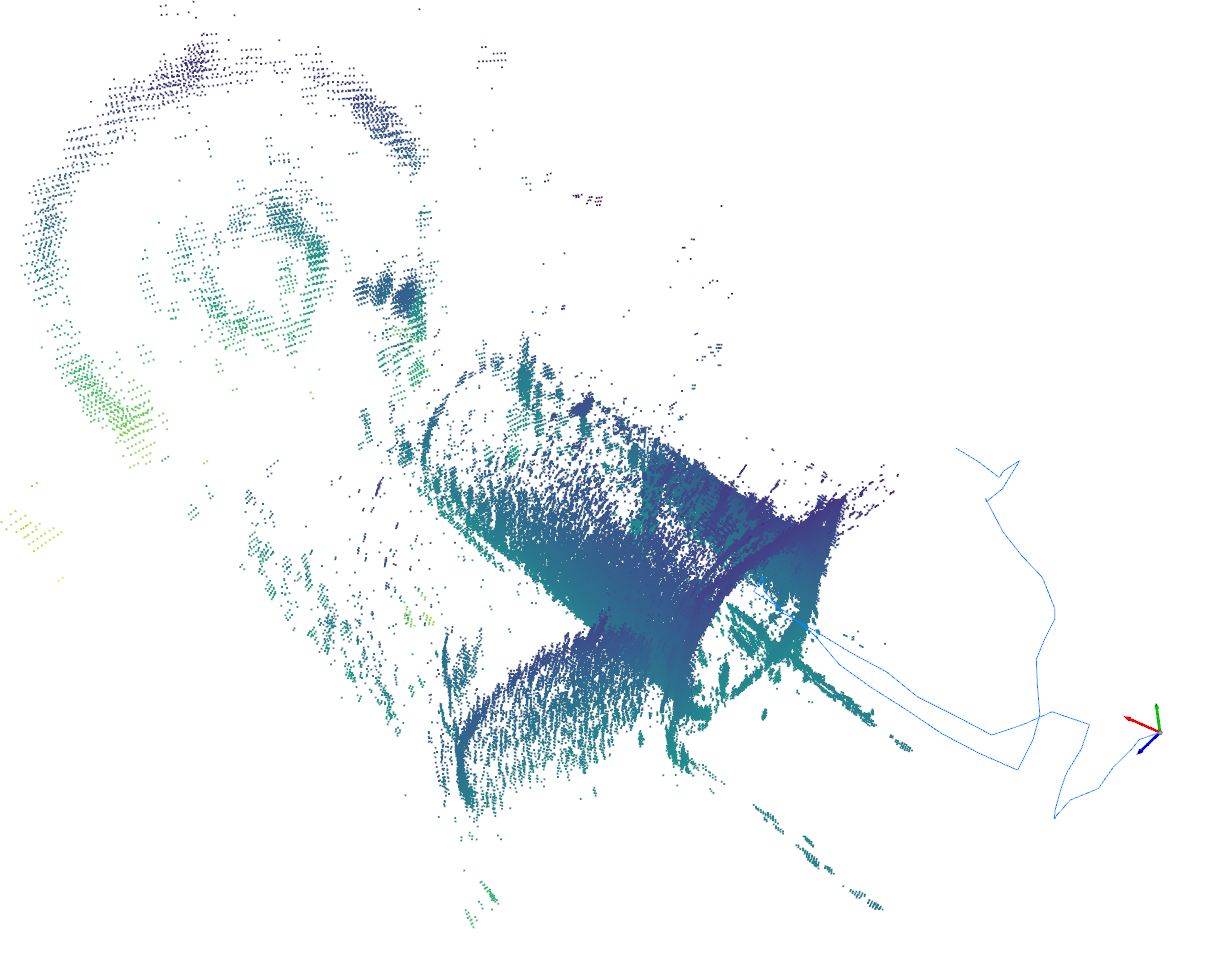}%
        \label{fig:pipe-container:c}%
    }

    \caption{Reconstruction outliers in acoustically challenging environments. The visibility conditions are shown in (a), a sonar scan registration in (b), with the the resulting reconstruction in (c). Outliers may be present due to multi-path issues.}
    \label{fig:pipe-container}
\end{figure}

To challenge the capacity of the sensor on proving valid range responses that could be utilized for reconstruction in confined spaces, a dataset was collected from a submerged steel cylindrical container, as shown in Fig.~\ref{fig:pipe-container} in poor visibility (Fig.~\ref{fig:pipe-container:a}).
In agreement to previous observations, it is possible that  certain sonar scans may be incomplete with some surfaces not returning a response such as the bottom of the door (Fig.~\ref{fig:pipe-container:b}), though details are covered with relative consistency such as the triangle features at the corners. 
In the full reconstruction (Fig.~\ref{fig:pipe-container:c}), the overall shape of the structure is captured well forming very accurately the cylindrical shape of the container, though several outliers are visible.

For instance, due to multipath issues when entering this confined space, inaccurate responses formed a weak, although noticeable outliers on the sides of the wall, at a distance similar to the diameter of the cylinder.
Moreover, a large interference pattern emerged in the form of two homocentric rings, on the other open-ended side of the cylindrical structure at a similar distance as its length, indicating further multipath inconsistencies.

\section{Lessons Learned, Future Work, and Conclusions}
\label{sec:conc}
One of the key observations was the effect of the field of view on the trajectory estimation. While the wider horizontal field of view ($90^\circ$) allowed tracking of features throughout the experiments, the restricted vertical field of view ($40$) during operations in the narrow tunnels of the Peacock cave system, FL, resulted in systematic drift over time. More specifically, as only a small patch of the floor and ceiling was visible for large segments of the trajectory, ICP resulted in incorrect matches and incremental drift on the Z-axis; loop-closures on the way back brought the trajectory back to the starting point. The final goal of this work will be to integrate all the available sources of information, acoustic, visual, inertial, water depth, and possibly magnetic field, into a tightly coupled optimization framework. 
Contrary to SVIn2~\cite{RahmanIJRR2022} where the pipe profiling sonar and the camera did not share a field of view, the Sonar 3D-15 shares the field of view of the camera, simplifying the correlation of features in the observed environment. 

The integration of the range measurements into a coherent volumetric map is also part of the future work. Such a development will assist in multi-session mapping where an environment is explored and mapped over several deployments. Commercial work by Sunfish Inc.~\cite{sunfish} has generated an example of visualization. Exploring different representations, such a voxels~\cite{siegel2023robotic}, Surflets~\cite{surflets}, or learning based Gaussian Splatting like representations in terms of accuracy and ease of integration into the state estimation process will open several directions of future research.

Finally, in challenging environments, as shown in our experiments, typical acoustic issues remain, such as multipath effects. Thus future research may focus on online corrections through probabilistic map representations.

Overall, this paper characterized a new 3D sonar in different challenging underwater environments for localization and dense reconstruction, providing the foundation for robust underwater mapping and exploration. By presenting a novel calibration strategy and SLAM pipelines and releasing our datasets, we aim to accelerate research into the unique challenges and opportunities of 3D acoustic sensing for enhanced underwater robot autonomy.

\section{Acknowledgments}
\label{sec:ackn}
The authors would like to thank  Water Linked AS for their unrestricted support, including access to their hardware and datasets during testing. Furthermore, we acknowledge the help from Woodville Karst Plain Project (WKPP), El Centro Investigador del Sistema Acuífero de Quintana Roo A.C. (CINDAQ), Global Underwater Explorers (GUE), Ricardo Constantino, and Project Baseline over the years in providing access to challenging underwater caves and mentoring us in underwater cave mapping. The authors are also grateful for equipment support by Halcyon Dive Systems, and KELDAN GmbH lights.

\bibliographystyle{IEEEtran}
\bibliography{./IEEEabrv,bibliography/refs}


\end{document}